\pdfoutput=1
\documentclass[11pt]{article}

\usepackage[final]{acl}

\usepackage{times}
\usepackage{latexsym}
\usepackage{caption}
\usepackage[T1]{fontenc}
\usepackage{enumitem}
\usepackage[utf8]{inputenc}

\usepackage{microtype}

\usepackage{inconsolata}

\usepackage{graphicx}
\usepackage{graphicx}

\usepackage[cmex10]{amsmath}
\usepackage{amssymb}
\usepackage{algorithmic}
\usepackage{algorithm}

\DeclareMathAlphabet\mathbfcal{OMS}{cmsy}{b}{n}
\newcommand{\ten}[1]{\mathbfcal{#1}}
\newcommand{\mat}[1]{\mathbf{#1}}

\title{Saten: Sparse Augmented Tensor Networks for Post-Training Compression of Large Language Models\thanks{Accepted to EMNLP 2025.}}

\author{
\textbf{Ryan Solgi\textsuperscript{1}, Kai Zhen\textsuperscript{2}, Rupak Vignesh Swaminathan\textsuperscript{2}, 
Nathan Susanj\textsuperscript{2},} \\ 
\textbf{Athanasios Mouchtaris\textsuperscript{2}, Siegfried Kunzmann\textsuperscript{2}, Zheng Zhang\textsuperscript{1}} \\
\textsuperscript{1}University of California-Santa Barbara, USA \\
\textsuperscript{2}Amazon, USA \\
\texttt{solgi@ucsb.edu}, \texttt{zhengzhang@ece.ucsb.edu}}

\begin{document}
\maketitle
\begin{abstract}

The efficient implementation of large language models (LLMs) is crucial for deployment on resource-constrained devices. Low-rank tensor compression techniques, such as tensor-train (TT) networks, have been widely studied for over-parameterized neural networks. However, their applications to compress pre-trained large language models (LLMs) for downstream tasks (post-training) remains challenging due to the high-rank nature of pre-trained LLMs and the lack of access to pretraining data. In this study, we investigate low-rank tensorized LLMs during fine-tuning and propose sparse augmented tensor networks (Saten) to enhance their performance. The proposed Saten framework enables full model compression. Experimental results demonstrate that Saten enhances both accuracy and compression efficiency in tensorized language models, achieving state-of-the-art performance.
\footnote{Representative code and implementation details are available at: \url{https://github.com/rmsolgi/saten.git}.}
\end{abstract}

\addtocounter{footnote}{-1}
\section{Introduction}
Transformers have shown great success in various natural language processing tasks~\citep{Vaswani-attention-2017, Devlin-bert-2019, Raffel-unified-2020}. However, their large number of parameters and computational demands hinder their implementation on resource-constrained devices. Consequently, various methods for LLM compression have been studied, including pruning~\citep{Sanh-pruning-2020}, distillation~\citep{Sanh-DistilBERT-2019}, quantization~\citep{Shen-qbert-2020}, and matrix factorization~\citep{Lan-albert-2020,Hsu-wsvd-2022, Gao-awsvd-2024}.

Low-rank tensor factorization is one of the prominent neural network compression techniques that has been widely studied and applied to various neural network architectures~\citep{Lebedev-cpcnn-2014,Yang-ttrnn-2017,hawkins-beysian-2019}. The tensor-train (TT) format is the most common tensor decomposition method used for neural network compression~\cite{Novikov-tnn-2015}. Furthermore, some studies successfully applied the idea of low-rank plus sparsity in tensor completion for signal processing~\citep{ mateos2012robust, driggs2019tensor}.

Despite their success in improving the efficiency of LLM fine-tuning and pre-training~\citep{Yang-adazeta-2024, Yang-loretta-2024, Yang-comera-2024}, low-rank tensor compression of pre-trained language models has not yet been successfully reported. This task mainly faces two challenges. First, almost all open-source LLMs are trained using uncompressed approaches, and the resulting model parameters of many layers may not exhibit a low-rank structure. Second, there is a lack of access to pre-training data to fully retrain low-rank parameters. These challenges result in a significant performance drop when post-training compression of language models is performed during fine-tuning using low-rank tensor factorization. 

\textbf{Paper Contributions.} In this study, we address the challenges mentioned above by introducing sparse augmented tensor networks (Saten). Our specific contributions are summarized below:
    \begin{itemize}[leftmargin=*]
    \vspace{-5pt}
        \item We propose Saten to compress LLMs with TT decomposition by incorporating a sparse error approximation. We study both unstructured and structured sparsity patterns in the Saten model. 
          \vspace{-5pt}
        \item We analyze the model and computational complexities of the model compressed by Saten. When the tensor ranks are low and sparsity ratios are high, our model shows both memory and computational savings in inference.   
          \vspace{-5pt}
        \item We compare Saten, TT and the recent SVD with adaptive rank selection~\cite{Gao-awsvd-2024} on BERT-Base and LlaMA. The experiments demonstrate that Saten achieves state-of-the-art performance in both compression efficiency and model accuracy. 
     
    \end{itemize}

\section{Background}
\label{sec: lrt-layer}

Tensors are generalizations of matrices to higher orders and can be represented by multi-dimensional data arrays \citep{tensor:suvey}. A real tensor of order $d$ and dimension can be denoted as $\ten{A} \in \mathbb{R}^{n_1\times n_2 \times \cdots \times n_d}$, where $n_j$ is the size of dimension $j$. The $(i_1, i_2, \cdots, i_d)$-th element of $\ten{A}$ is $a_{i_1 i_2 \cdots i_d}$. Obviously, matrices and vectors are tensors of order $d=2$ and $d=1$, respectively.

\paragraph{Tensor Train (TT) Decomposition.} Many practical tensors have low ranks and can be approximated using low-rank tensor decompositions. The TT decomposition approximates an order-$d$ tensor $\ten{A}$ with $d$ factors 
$\{ \ten{G}_j \in \mathbb{R}^{r_{j-1}\times n_j \times r_{j}} \}_{j=1}^d$:
\begin{align} 
\label{eq:TT_format}
\ten{A}=\ten{G}_1\times_{3}^{1}\ten{G}_2\times_{3}^{1}\cdots\times_{3}^{1}\ten{G}_d,
\end{align}
where $r_0, r_1,\cdots, r_d$ are TT ranks and $r_0=r_d=1$. Here, operator $\times_{3}^{1}$ refers to contracting the third and first dimensions of the left and right tensors, respectively (see Appendix~\ref{app:tt} for more details).

\section{The Saten Method}
In this section, we present the Saten framework to compress a pre-trained LLM. Since many layers of a pre-trained LLM may not exhibit a low-rank property, compressing a pre-trained model directly using TT decomposition can result in huge accuracy drop. To address this challenge, Saten approximates the matrix $\mat{W}$ as follows:

\begin{align}
    \label{eq:saten}
    \mat{W}\approx \hat{\mat{W}}_{\text{TT}}+ {\mat{E}}
\end{align}
where ${\mat{E}}$ is a sparse matrix, $\hat{\mat{W}}_{\text{TT}}$ is the matrixization of a tensor $\hat{\ten{W}}_{\text{TT}}$ with a TT representation: 
\begin{equation}
\label{eq:TT_part}
    \hat{\ten{W}}_{\text{TT}}=\ten{G}_1 \times_{3}^1 \ten{G}_2 \times_{3}^1  \cdots \times_{3}^1 \ten{G}_{d+k}.
\end{equation}

\begin{figure}[t]
\centering
\includegraphics[width=3.1in]{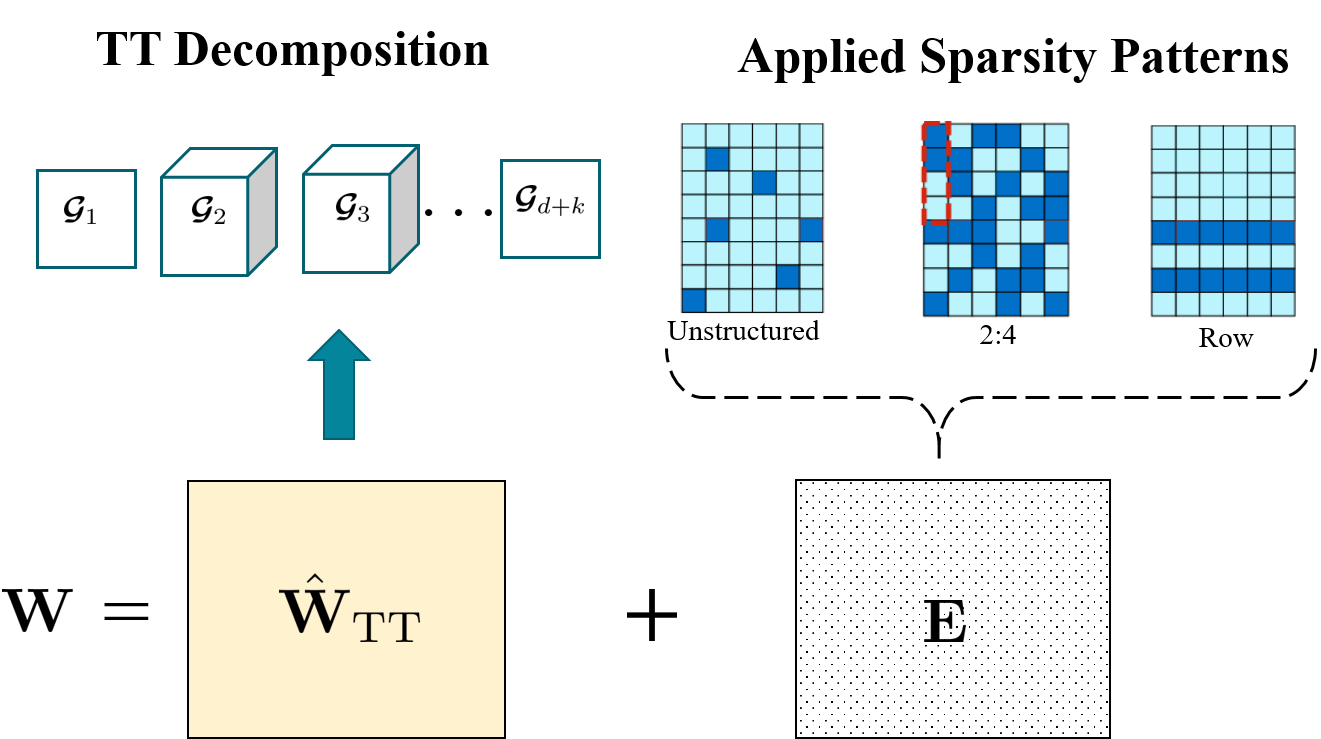}
\caption{The sparse + low-rank tensor train representation for a weight matrix (or embedding table).}
\label{fig:saten}
\end{figure}

The decomposition in Saten is illustrated in Fig.~\ref{fig:saten}.

\subsection{Computing the Saten Model}

\paragraph{Low-rank TT component.} We fist get the low-rank TT repsentation via the following steps:
\begin{itemize}[leftmargin=*]
    \item {\bf Step 1:} We factor the dimensions of the matrix $\mat{W}$ as $N=n_1\times n_2 \cdots \times n_k$ and $M=m_1\times m_2 \cdots \times m_d$, and fold $\mat{W}$ to an order-$(d+k)$ tensor $\ten{W} \in \mathbb{R}^{m_1 \times \cdots m_d \times n_1 \cdots n_k}$.
    \item {\bf Step 2:} We perform TT decomposition over $\ten{W}$:
    \begin{equation}
    \label{eq:TT_comp}
   \ten{W} \approx  \hat{\ten{W}}_{\text{TT}}=\ten{G}_1 \times_{3}^1 \ten{G}_2 \times_{3}^1  \cdots \times_{3}^1 \ten{G}_{d+k}.
\end{equation}
\end{itemize}
The accuracy of the TT decomposition highly depends on two factors: the TT ranks $(r_0, r_1 \cdots r_{d+k})$, and tensor shape $(n_1, \cdots n_d, m_1, \cdots m_k)$. In Saten, we apply an error-based TT-SVD method (see Appendix~\ref{sec:tensor_train}) in each layer to automatically determine the TT ranks, which greatly reduces the number of hyperparameters. In order to find a memory-efficient tensor shape in the compression process, Saten solves the tensor shape optimization problem using the method described in Appendix~\ref{sec:tensor-shape}.

\paragraph{Sparse component.} After getting the TT factors in \eqref{eq:TT_part}, we further determine the sparse matrix $\mat{E}$. We assume that $\mat{M}\in\{0,1\}^{N\times M}$ is a sparsity mask for $\mat{E}$, and the sparse matrix is decided as:
\begin{equation}
    \mat{E}=\left(\mat{W}-\hat{\mat{W} }_{\text{TT}}\right) \odot \mat{M},
\end{equation}
where $\odot$ denotes element-wise product operations. To compress the linear layers, we have implemented two different methods: Saten(u) and Saten(2:4). 
\begin{itemize}[leftmargin=*]
    \item {\bf Saten(u).} In this method, we consider unstructured sparsity: the error is pruned (by setting the associated elements in $\mat{M}$ as zero) to preserve the top percent absolute values of matrix $\mat{W}-\hat{\mat{W}}_{\text{TT}}$.
    \item {\bf Saten(2:4).} In this method we consider the 2:4 structured sparsity to enhance the efficiency on GPU. Specifically, of every $4$ elements along a column of the error matrix $\mat{W}-\hat{\mat{W}}_{\text{TT}}$, the $2$ largest-magnitude elements are preserved in $\mat{E}$.
\end{itemize} 
When $\mat{W}$ is the parameter in an embedding layer, a different sparsity pattern ({\bf row sparsity}) is used to compute $\mat{E}$. Specifically, the error $\mat{W}-\hat{\mat{W}}_{\text{TT}}$ is pruned to retain only the rows corresponding to the most frequent tokens in the training dataset.

\paragraph{Fine-tuning.} After deciding the values of low-rank TT factors and the sparse matrix, we still need to fine-tune the compressed model to further boost the performance. This can be done directly in the {\bf compressed format}. Specifically, the matrix $\mat{W}$ does not need to be recovered, and we can use automatic differentiation to compute directly the stochastic gradient w.r.t. the TT factors $\{\ten{G}_j\}_{j=1}^{d+k}$ and w.r.t. the sparse matrix $\mat{E}$. Then these compressed components are updated via stochastic gradient descent. 

The overall flow of the Saten framework is summarized in Algorithm~\ref{alg:saten}.

\begin{algorithm}[t]
\caption{Language Model Compression using Saten}
\label{alg:saten}
\begin{algorithmic}
\REQUIRE{A pre-trained model} 
\FOR{layer \textbf{in} targeted layers}
\STATE Extract layer's weight ($\mat{W}$);
\STATE Fold $\mat{W}$ into a high-order tensor $\ten{W}$;
\STATE Compute TT factors for $\ten{W}$ [ Eq.~\eqref{eq:TT_comp}]
\STATE Compute sparse error; $\mat{E}$ with a chosen sparsity pattern;
\STATE Replace layer by the Saten layer [Eq.~\eqref{eq:saten_linear}].
\ENDFOR
\STATE Fine-tune the Saten model.
\STATE \textbf{Return}  the Saten model
\end{algorithmic}
\end{algorithm}

\begin{table*}[t]
\small
\caption{The experimental results of the end-to-end compression of BERT for GLUE datasets (SVD and SVD-ARS are cited from~\citet{Gao-awsvd-2024}). (For SST-2, MNLI, and QNLI, the values represent accuracy. For MRPC and QQP, they correspond to the F1 score. Values for CoLA and STS-B refer to Matthew’s correlation and Pearson correlation, respectively). }
\label{tab:bert-glue}
\begin{center}
\begin{tabular}{c|cccccccc}
\multicolumn{1}{c}{Model}
&\multicolumn{1}{c}{MRPC}
&\multicolumn{1}{c}{STSB}
&\multicolumn{1}{c}{CoLA}
&\multicolumn{1}{c}{SST-2}
&\multicolumn{1}{c}{MNLI}
&\multicolumn{1}{c}{QNLI}
&\multicolumn{1}{c}{QQP}
&\multicolumn{1}{c}{\# Params (M)}
\\ \hline
BERT-Base&91.50&89.42&58.92&92.78&84.31&91.38&87.95&109.5\\
\hline
Saten(2:4)&\textbf{90.19}&\textbf{87.23}&45.28&91.86&81.74&\textbf{89.82}&86.87&\textbf{59.9}\\
TT&84.57&84.32&10.25&89.10&79.02&87.71&86.33&64.8\\
SVD (\citet{Gao-awsvd-2024})&83.60&85.67&29.02&91.28&83.02&89.35&87.05&66.5\\
SVD-ARS (\citet{Gao-awsvd-2024})&85.57&86.30&\textbf{47.08}&\textbf{91.97}&\textbf{83.55}&89.44&\textbf{87.39}&65.1\\

\hline
Saten(u)&\textbf{83.38}&\textbf{83.43}&23.60&\textbf{90.25}&80.87&\textbf{87.61}&\textbf{86.91}&\textbf{50.3}\\
SVD (\citet{Gao-awsvd-2024})&81.06&79.35&9.83&89.11&81.61&86.99&86.35&52.4\\
SVD-ARS (\citet{Gao-awsvd-2024})&81.42&82.85&\textbf{27.62}&89.22&\textbf{83.07}&87.50&86.68&52.6\\
\hline
\end{tabular}
\end{center}
\end{table*}

\begin{table*}[t]
\small
\caption{The experimental results of the compression of LlaMA-3.2-1B using Saten(2:4), Saten(u), and TT. } 
\label{tab:llama}
\begin{center}
\begin{tabular}{ccc|cccccc}
\multicolumn{1}{c}{Model}
&\multicolumn{1}{c}{$L_{\text{lin}}$(\%)}
&\multicolumn{1}{c}{$\rho_{\text{lin}}$(\%)}
&\multicolumn{1}{c}{BoolQ(\%)}
&\multicolumn{1}{c}{CB(\%)}
&\multicolumn{1}{c}{WSC(\%)}
&\multicolumn{1}{c}{COPA(\%)}
&\multicolumn{1}{c}{\#Params (B)}
&\multicolumn{1}{c}{\#MACs (M)}
\\ \hline
Llama-3.2-1B&-&-&66.48&87.50&63.46&65.00&1.24&973\\
\hline

Saten(2:4)&13&50&66.29&\textbf{91.07}&\textbf{64.42}&\textbf{65.00}&0.78&630\\
TT&68&0&64.83&58.93&50.00&55.00&0.93&706\\
Tucker&71&0&63.33&66.07&63.46&47.00&0.95&744\\
SVD&68&-&65.66&69.64&63.46&51.00&0.93&615\\
\hline
Saten(u)&16&5&\textbf{67.03}&71.43&57.69&52.00&\textbf{0.38}&\textbf{226}\\
TT&27&0&62.08&51.79&42.31&51.00&0.53&293\\
Tucker&29&0&63.94&66.07&55.76&49.00&0.55&320\\
SVD&27&-&65.50&67.86&61.54&50.00&0.53&270\\
\hline

\hline
\end{tabular}
\end{center}
\end{table*}

\subsection{Complexity of Saten} 
In this subsection, we analyze the model complexity and computational complexity of the compressed model by Saten.

\paragraph{Number of Parameters:} The number of model parameters of a Saten representation of a linear transformation is computed as follows:
\begin{align}
    P=\sum_{j=1}^{d+k}r_{j-1}s_jr_j + \rho NM.
\end{align}
where $\mat{s}=(n_1,\cdots,n_k,m_1,\cdots,m_d)$, and $\rho$ is the density of non-zero elements in $\mat{E}$. When the TT ranks and sparsity are sufficiently small, $P$ becomes smaller than the number of parameters in a linear layer ($NM$).

\paragraph{Computational Complexities:} Let $\mat{x} \in \mathbb{R}^{N}$ be the input of a linear layer $\mat{y}=\mat{W}^T\mat{x}$. This linear layer can be written as $\mat{y}=\hat{\mat{W}}_{\text{TT}}^T\mat{x} +\mat{E}^T\mat{x}$ in the matrix format after applying the Saten compression flow. However, since $\hat{\mat{W}}_{\text{TT}}$ is parameterized in a TT format, we rewrite the linear layer by using the TT factors directly, leading to
\begin{align}
\label{eq:saten_linear}
    \mat{y}={\mat{E}}^T\mat{x}+f(\ten{X},\ten{G}_1,\cdots,\ten{G}_{d+k}).
\end{align}
Here $\ten{X} \in \mathbb{R}^{n_1\times\cdots\times n_k}$ is obaianed by folding $\mat{x}$ into an order-$k$ tensor, $f$ denotes the process of contracting $\ten{X}$ with the TT factors as described in Appendix~\ref{app:tt}. This tensor network contraction leads to the following number of MACs (multiplication and accumulation)
\begin{align}
\label{eq:mac-tt}
C_{\text{TT}}&=\sum_{i=1}^{k} \frac{Nr_{i-1}r_{i}}{\prod_{j=1}^{i-1}n_j} + \sum_{i=k+1}^{d}(\prod_{j=k+1}^{i}m_{j}) r_{i-1}r_{i},
\end{align}
which is usually much smaller than that of a standard matrix-vector multiplication. The total number MACs for the inference of a Saten layer is
\begin{align}
\label{eq:mac-saten}
C_{\text{Saten}}=C_{\text{TT}}+(\rho N + 1)M.
\end{align}
When the TT ranks and $\rho$ are sufficiently small, $C_{\text{Saten}}$ becomes smaller than the typical linear layer MAC count, which is equal to $M(N+1)$.

\section{Experimental Results}
In this section, we conduct experiments to compress BERT-Base~\citep{Devlin-bert-2019} and LlaMA-3.2-1B~\citep{meta2024llama}. For details regarding hyperparameter for fine-tuning and Saten compression, see Appendix~\ref{app:run-details}.  In Appendix~\ref{app:DistilBert}, we also show additional results regarding the compression of DistilBERT~\cite{Sanh-DistilBERT-2019}, with a particular focus on (1) the effectiveness of Saten on the embedding layer, (2) the balance between sparsity and low-rank parameterization.  In the following, we focus on discussing the overall performance on BERT-Base and LlaMA-3.2-1B.

\subsection{BERT Compression for GLUE}
\label{sec:bert-base}
Table~\ref{tab:bert-glue} compares Saten with other compression methods, including TT, singular value decomposition (SVD), and the most recent SVD with adaptive rank selection (SVD-ARS)~\cite{Gao-awsvd-2024}, on the GLUE benchmarks~\citep{Wang-GLUE-2018} for BERT-Base. It can be seen that Saten achieves the {\bf highest compression ratio} (by nearly a factor of two) while preserving the performance of the network compared to the BERT-Base model. Meanwhile, Saten produces much {\bf better accuracy} compared to SVD and TT. Although Saten is not equipped with adaptive rank selection, it still achieves {\bf competitive accuracy with fewer model parameters} compared to SVD-ARS, which optimizes SVD ranks. 

Both Saten(2:4) and Saten(u) have 8.8 million parameters in the word embedding layer. Table~\ref{tab:spec-bert} presents details of the encoder (linear layers) compressed by Saten, including their low-rank and sparsity budgets and MAC count relative to the base model. $\rho_{\text{lin}}$ and $L_{\text{lin}}$ denote the proportion of non-zero elements and low-rank parameters relative to the original total number of parameters in the linear layers, respectively. In Saten(u), linear layers have 95\% sparsity and 40\% low-rank parameters, leading to $1.9\times$ reduction in the network's MAC count. Meanwhile, Saten(2:4), which achieves performance comparable to the base model, reduces the MAC count by $1.7\times$.

\subsection{LLaMA Compression}
We further apply Saten  to compress LLaMA-3.2-1B across multiple datasets from SuperGLUE benchmarks~\citep{wang2019superglue} 

%including CB~\citep{demarneffe:cb}, WSC~\citep{levesque2011winograd}, COPA~\citep{copaRoemmele}, and BoolQ~\citep{clark2019boolq}. 

Table~\ref{tab:llama} presents the specifications and results of Saten(u) and Saten(2:4), compared with TT, Tucker, and SVD. Both Saten(2:4) and Saten(u) have 170 million low-rank parameters with 92\% sparsity in their embedding layer. Saten(2:4) outperforms the base model on CB and WSC while reducing model size by about 40\% and maintaining accuracy on the BoolQ dataset. Its improved accuracy stems from better generalization and reduced overfitting. Meanwhile, Saten(u) achieves over 3× and 4× reductions in model size and MAC counts, respectively.

\begin{table}[t]
\small
\caption{Specification of the Saten layers relative to the BERT-Base linear layers for GLUE benchmarks.} 
\label{tab:spec-bert}
\begin{center}
\begin{tabular}{c|ccccc}
\multicolumn{1}{c}{Model}
&\multicolumn{1}{c}{$L_{\text{lin}}(\%)$}
&\multicolumn{1}{c}{$\rho_{\text{lin}}(\%)$}
&\multicolumn{1}{c}{MAC Reduction}
\\ \hline
Saten(u)&40&5&1.9$\times$\\
Saten(2:4)&8&50&1.7$\times$&\\
\hline
\end{tabular}
\end{center}
% \vspace{-10pt}
\end{table}

\section{Conclusion}
Pre-trained model parameters may exist high-rank properties in many layers, leading to performance drops when using tensor factorization for LLM post-training compression. We have presented Saten, which integrates a sparse component into tensor networks to improve model performance in compression. Saten has been tested on BERT-Base and LlaMA-3.2-1B, showing best compression ratios and sate-of-the-art accuracy compared to SVD, TT and the most recent SVD-ARS.

\section{Limitations}

We demonstrated that Saten reduces the theoretical computational time complexities (i.e. MACs) compared to the uncompressed model. Current mainstream platforms for implementing large language models are not optimized for TT plus sparse operations. Nevertheless, there is ongoing interest and progress in improving sparse operations for AI and machine learning tasks~\citep{Fu-JITSPMM-2024, Hsu-sam-2023}. Meanwhile, in this study, we utilized structured sparsity, which facilitates the implementation of Saten on GPUs~\citep{Zhou-m2n-2021} and kept the unstructured sparsity at high levels of 95\% for which sparse algebra packages typically achieve significant speedup. Our findings motivate future research on custom hardware accelerators as the actual inference speedup of Saten requires both software- and hardware-level optimization, which is the subject of our future studies. Another promising direction for future work is the application of quantization to further enhance compression efficiency and reduce inference cost~\citep{saha2024compressing}. Finally, we evaluated the LLaMA 3.2 models, which are already distilled and thus more difficult to compress. This highlights Saten’s practical utility in resource-constrained, low-redundancy settings. Although we do not include further larger-scale models, Saten’s architecture-agnostic design and layer-wise compression make it scalable, and we expect the observed improvements to generalize across model sizes and architectures.

\bibliography{main}

\begin{thebibliography}{34}
\providecommand{\natexlab}[1]{#1}

\bibitem[{AI(2024)}]{meta2024llama}
Meta AI. 2024.
\newblock Llama 3.2-1b model card.
\newblock Retrieved from \url{https://huggingface.co/meta-llama/Llama-3.2-1B}.

\bibitem[{Devlin et~al.(2019)Devlin, Chang, Lee, and Toutanova}]{Devlin-bert-2019}
Jacob Devlin, Ming-Wei Chang, Kenton Lee, and Kristina Toutanova. 2019.
\newblock Bert: Pre-training of deep bidirectional transformers for language understanding.
\newblock \emph{Proceedings of the NAACL-HLT}.

\bibitem[{Driggs et~al.(2019)Driggs, Becker, and Boyd-Graber}]{driggs2019tensor}
Derek Driggs, Stephen Becker, and Jordan Boyd-Graber. 2019.
\newblock Tensor robust principal component analysis: Better recovery with atomic norm regularization.
\newblock \emph{arXiv preprint arXiv:1901.10991}.

\bibitem[{Fu et~al.(2024)Fu, Rolinger, and Huang}]{Fu-JITSPMM-2024}
Qiang Fu, Thomas~B. Rolinger, and H.~Howie Huang. 2024.
\newblock Jitspmm: Just-in-time instruction generation for accelerated sparse matrix-matrix multiplication.
\newblock \emph{Proceedings of the 2024 IEEE/ACM International Symposium on Code Generation and Optimization}.

\bibitem[{Gao et~al.(2024)Gao, Hua, Hsu, Shen, and Jin}]{Gao-awsvd-2024}
Shangqian Gao, Ting Hua, Yen-Chang Hsu, Yilin Shen, and Hongxia Jin. 2024.
\newblock Adaptive rank selections for low-rank approximation of language models.
\newblock \emph{Proceedings of the 2024 Conference of the North American Chapter of the Association for Computational Linguistics}.

\bibitem[{Hawkins and Zhang(2019)}]{hawkins-beysian-2019}
Cole Hawkins and Zheng Zhang. 2019.
\newblock Bayesian tensorized neural networks with automatic rank selection.
\newblock \emph{arXiv:1905.10478}.

\bibitem[{Hsu et~al.(2023)Hsu, Strange, Sharma, Won, Olukotun, Emer, Horowitz, and Kjølstad}]{Hsu-sam-2023}
Olivia Hsu, Maxwell Strange, Ritvik Sharma, Jaeyeon Won, Kunle Olukotun, Joel~S Emer, Mark~A Horowitz, and Fredrik Kjølstad. 2023.
\newblock The sparse abstract machine.
\newblock \emph{Proceedings of the 28th ACM International Conference on Architectural Support for Programming Languages and Operating Systems,}.

\bibitem[{Hsu et~al.(2022)Hsu, Hua, Chang, Lou, Shen, and Jin}]{Hsu-wsvd-2022}
Yen-Chang Hsu, Ting Hua, Sungen Chang, Qian Lou, Yilin Shen, and Hongxia Jin. 2022.
\newblock Language model compression with weighted low-rank factorization.
\newblock \emph{International Conference on Learning Representations (ICLR)}.

\bibitem[{Hu et~al.(2021)Hu, Shen, Wallis, Allen-Zhu, Li, Wang, and Chen}]{hu2021lora}
Edward~J. Hu, Yelong Shen, Phillip Wallis, Zeyuan Allen-Zhu, Yuanzhi Li, Shean Wang, and Weizhu Chen. 2021.
\newblock \href {https://arxiv.org/abs/2106.09685} {Lora: Low-rank adaptation of large language models}.
\newblock \emph{arXiv preprint arXiv:2106.09685}.

\bibitem[{Kolda and Bader(2009)}]{tensor:suvey}
Tamara~G. Kolda and Brett~W. Bader. 2009.
\newblock Tensor decompositions and applications.
\newblock \emph{SIAM Review}, 51(3):455--500.

\bibitem[{Lan et~al.(2020)Lan, Chen, Goodman, Gimpel, Sharma, and Soricut}]{Lan-albert-2020}
Zhenzhong Lan, Mingda Chen, Sebastian Goodman, Kevin Gimpel, Piyush Sharma, and Radu Soricut. 2020.
\newblock Albert: A lite bert for self-supervised learning of language representations.
\newblock \emph{International Conference on Learning Representations (ICLR)}.

\bibitem[{Lebedev et~al.(2014)Lebedev, Ganin, Rakhuba, Oseledets, and Lempitsky}]{Lebedev-cpcnn-2014}
Vadim Lebedev, Yaroslav Ganin, Maksim Rakhuba, Ivan Oseledets, and Victor Lempitsky. 2014.
\newblock Speeding-up convolutional neural networks using fine-tuned cp-decomposition.
\newblock \emph{arXiv preprint arXiv:1412.6553}.

\bibitem[{Mateos and Giannakis(2012)}]{mateos2012robust}
Gonzalo Mateos and Georgios~B. Giannakis. 2012.
\newblock Robust pca as bilinear decomposition with outlier-sparsity regularization.
\newblock \emph{IEEE Transactions on Signal Processing}, 60(10):5176--5190.

\bibitem[{Merity et~al.(2016)Merity, Xiong, Bradbury, and Socher}]{wikitext}
Stephen Merity, Caiming Xiong, James Bradbury, and Richard Socher. 2016.
\newblock Pointer sentinel mixture models.
\newblock \emph{arXiv preprint arXiv:1609.07843}.

\bibitem[{Mu et~al.(2013)Mu, Huang, Wright, and Goldfarb}]{mu2013square}
C.~Mu, B.~Huang, J.~Wright, and D.~Goldfarb. 2013.
\newblock Square deal: Lower bounds and improved relaxations for tensor recovery.
\newblock \emph{arXiv:1307.5870v2}.

\bibitem[{Novikov et~al.(2015)Novikov, Podoprikhin, Osokin, and Vetrov}]{Novikov-tnn-2015}
Alexander Novikov, Dmitrii Podoprikhin, Anton Osokin, and Dmitry~P Vetrov. 2015.
\newblock Tensorizing neural networks.
\newblock \emph{Advances in neural information processing systems}.

\bibitem[{Oseledets(2011)}]{Oseledets-tt-2011}
Ivan Oseledets. 2011.
\newblock Tensor-train decomposition.
\newblock \emph{SIAM Journal on Scientific Computing}.

\bibitem[{Raffel et~al.(2020)Raffel, Shazeer, Roberts, Lee, Narang, Matena, Zhou, Li, and Liu}]{Raffel-unified-2020}
Colin Raffel, Noam Shazeer, Adam Roberts, Katherine Lee, Sharan Narang, Michael Matena, Yanqi Zhou, Wei Li, and Peter~J. Liu. 2020.
\newblock Exploring the limits of transfer learning with a unified text-to-text transformer.
\newblock \emph{Journal of Machine Learning Research (JMLR)}.

\bibitem[{Saha et~al.(2024)Saha, Sagan, Srivastava, Goldsmith, and Pilanci}]{saha2024compressing}
Rajarshi Saha, Naomi Sagan, Varun Srivastava, Andrea Goldsmith, and Mert Pilanci. 2024.
\newblock Compressing large language models using low rank and low precision decomposition.
\newblock In \emph{Advances in Neural Information Processing Systems}, volume~37, pages 88981--89018.

\bibitem[{Sanh et~al.(2019)Sanh, Debut, Chaumond, and Wolf}]{Sanh-DistilBERT-2019}
Victor Sanh, Lysandre Debut, Julien Chaumond, and Thomas Wolf. 2019.
\newblock Distilbert, a distilled version of bert: Smaller, faster, cheaper, and lighter.
\newblock \emph{arXiv:1910.01108}.

\bibitem[{Sanh et~al.(2020)Sanh, Wolf, and Rush}]{Sanh-pruning-2020}
Victor Sanh, Thomas Wolf, and Alexander~M. Rush. 2020.
\newblock Movement pruning: Adaptive sparsity by fine-tuning.
\newblock \emph{Advances in Neural Information Processing Systems (NeurIPS).}

\bibitem[{Shen et~al.(2020)Shen, Dong, Ye, Ma, Yao, Gholami, Mahoney, and Keutzer}]{Shen-qbert-2020}
Sheng Shen, Zhen Dong, Jiayu Ye, Linjian Ma, Zhewei Yao, Amir Gholami, Michael~W. Mahoney, and Kurt Keutzer. 2020.
\newblock Q-bert: Hessian based ultra low precision quantization of bert.
\newblock \emph{AAAI Conference on Artificial Intelligence.}

\bibitem[{Solgi(2024)}]{solgi-thesis-2024}
Ryan Solgi. 2024.
\newblock Low-rank tensorized neural networks with tensor geometry optimization.
\newblock Master's thesis, Department of Electrical and Computer Engineering, University of California Santa Barbra.

\bibitem[{Solgi et~al.(2023)Solgi, He, Liang, Zhang, and Loaiciga}]{solgi-ts-2023}
Ryan Solgi, Zichang He, William~Jiahua Liang, Zheng Zhang, and Hugo~A Loaiciga. 2023.
\newblock Tensor shape search for efficient compression of tensorized data and neural networks.
\newblock \emph{Applied Soft Computing}.

\bibitem[{Vaswani et~al.(2017)Vaswani, Shazeer, Parmar, Uszkoreit, Llion~Jones, Kaiser, and Polosukhin}]{Vaswani-attention-2017}
Ashish Vaswani, Noam Shazeer, Niki Parmar, Jakob Uszkoreit, Aidan N.~Gomez Llion~Jones, Lukasz Kaiser, and Illia Polosukhin. 2017.
\newblock Attention is all you need.
\newblock \emph{Advances in Neural Information Processing Systems (NeurIPS)}.

\bibitem[{Wang et~al.(2019)Wang, Pruksachatkun, Nangia, Singh, Michael, Hill, Levy, and Bowman}]{wang2019superglue}
Alex Wang, Yada Pruksachatkun, Nikita Nangia, Amanpreet Singh, Julian Michael, Felix Hill, Omer Levy, and Samuel~R. Bowman. 2019.
\newblock \href {https://papers.nips.cc/paper/2019/hash/4496bf24afe7fab6f046bf4923da8de6-Abstract.html} {Superglue: A stickier benchmark for general-purpose language understanding systems}.
\newblock In \emph{Advances in Neural Information Processing Systems}, volume~32.

\bibitem[{Wang et~al.(2018)Wang, Singh, Michael, Hill, Levy, and Bowman}]{Wang-GLUE-2018}
Alex Wang, Amanpreet Singh, Julian Michael, Felix Hill, Omer Levy, and Samuel~R. Bowman. 2018.
\newblock Glue: A multi-task benchmark and analysis platform for natural language understanding.
\newblock \emph{Association for Computational Linguistics}.

\bibitem[{Wolf et~al.(2020)Wolf, Debut, Sanh, Chaumond, Delangue, Moi, Cistac, Rault, Louf, Funtowicz, Davison, Shleifer, von Platen, Ma, Jernite, Plu, Xu, Le~Scao, Gugger, Drame, Lhoest, and Rush}]{wolf-transformers-2020}
Thomas Wolf, Lysandre Debut, Victor Sanh, Julien Chaumond, Clement Delangue, Anthony Moi, Pierric Cistac, Tim Rault, Remi Louf, Morgan Funtowicz, Joe Davison, Sam Shleifer, Patrick von Platen, Clara Ma, Yacine Jernite, Julien Plu, Canwen Xu, Teven Le~Scao, Sylvain Gugger, Mariama Drame, Quentin Lhoest, and Alexander~M. Rush. 2020.
\newblock \href {https://doi.org/10.18653/v1/2020.emnlp-demos.6} {Transformers: State-of-the-art natural language processing}.
\newblock In \emph{Proceedings of the 2020 Conference on Empirical Methods in Natural Language Processing: System Demonstrations}, pages 38--45, Online. Association for Computational Linguistics.

\bibitem[{Yang et~al.(2024{\natexlab{a}})Yang, Zhen, Banijamal, Mouchtaris, and Zhang}]{Yang-adazeta-2024}
Yifan Yang, Kai Zhen, Ershad Banijamal, Athanasios Mouchtaris, and Zheng Zhang. 2024{\natexlab{a}}.
\newblock Adazeta: Adaptive zeroth-order tensor-train adaption for memory-efficient large language models fine-tuning.
\newblock \emph{Conference on Empirical Methods in Natural Language Processing (EMNLP)}.

\bibitem[{Yang et~al.(2024{\natexlab{b}})Yang, Zhou, Wong, and Zhang}]{Yang-loretta-2024}
Yifan Yang, Jiajun Zhou, Ngai Wong, and Zheng Zhang. 2024{\natexlab{b}}.
\newblock Loretta: Low-rank economic tensor-train adaptation for ultra-low-parameter fine-tuning of large language models.
\newblock \emph{Conf. Northern American Association of Computational Linguistics (NAACL)}.

\bibitem[{Yang et~al.(2017)Yang, Krompass, and Tresp}]{Yang-ttrnn-2017}
Yinchong Yang, Denis Krompass, and Volker Tresp. 2017.
\newblock Tensor train recurrent neural networks for video classification.
\newblock \emph{Proceedings of the British Machine Vision Conference (BMVC)}.

\bibitem[{Yang et~al.(2024{\natexlab{c}})Yang, Liu, Choudhary, Xie, Gao, Kunzmann, and Zhang}]{Yang-comera-2024}
Zi~Yang, Ziyue Liu, Samridhi Choudhary, Xinfeng Xie, Cao Gao, Siegfried Kunzmann, and Zheng Zhang. 2024{\natexlab{c}}.
\newblock Comera: Computing- and memory-efficient training via rank-adaptive tensor optimization.
\newblock \emph{arXiv:2405.14377}.

\bibitem[{Zhong et~al.(2019)Zhong, Wei, Lin, and Zhang}]{Zhong2019}
Z.~Zhong, F.~Wei, Z.~Lin, and C.~Zhang. 2019.
\newblock Ada-tucker: Compressing deep neural networks via adaptive dimension adjustment tucker decomposition.
\newblock \emph{arXiv:1906.07671}.

\bibitem[{Zhou et~al.(2021)Zhou, Ma, Zhu, Liu, Zhang, Yuan, Sun, and Li}]{Zhou-m2n-2021}
Aojun Zhou, Yukun Ma, Junnan Zhu, Jianbo Liu, Zhijie Zhang, Kun Yuan, Wenxiu Sun, and Hongsheng Li. 2021.
\newblock Learning n:m fine-grained structured sparse neural networks from scratch.
\newblock \emph{arXiv:2102.04010}.

\end{thebibliography}

\appendix
\section{Tensor Train Network}
\label{app:tt}

\paragraph{Tensor:} A tensor is a generalization of scalars, vectors, and matrices to higher dimensions. A tensor of order $d$ over the real numbers is an array of components indexed by $d$ indices as follows:
\begin{align}
    &\ten{A} = \big( a_{i_1,i_2,\dots,i_d} \big) \in \mathbb{R}^{I_1 \times I_2 \times \cdots \times I_d}, \nonumber \\
    &\forall i_t = 1,2,\dots,I_t, \quad t = 1,2,\dots,d.
\end{align}

\paragraph{Tensor Contraction:}
A tensor contraction generalizes matrix multiplication to tensors. Without loss of generality, the contraction of two tensors $\ten{A}\in \mathbb{R}^{I_1 \times \cdots \times I_{v}\times T_1 \times \cdots \times T_h}$ and $\ten{B}\in \mathbb{R}^{T_1 \times \cdots \times T_{h}\times J_1 \times \cdots \times J_u}$ over the shared dimensions $T_1,\cdots,T_h$ results in a new tensor $\ten{C}\in \mathbb{R}^{I_1\times \cdots \times I_v \times J_1 \times \cdots \times J_u}$ with elements given by:

\begin{align}
\label{eq:contraction}
&c_{i_1,\cdots,i_v,j_1,\cdots,j_u}\nonumber\\ 
&=\sum_{t_1,\cdots,t_h} a_{i_1,\cdots,i_v,t_1,\cdots,t_h} b_{t_1,\cdots,t_h,j_1,\cdots,j_u}, \nonumber\\
&\forall \quad i_1,\cdots,i_v, \quad j_1,\cdots,j_u.
\end{align}
For arbitrary tensors $\ten{A}$ and $\ten{B}$ we define a contraction pattern as a set of index pairs $\{(l_1,l'_1),\cdots,(l_h,l'_h)\}$ where the indices $l_i$ in $\ten{A}$ correspond to indices $l'_i$ in $\ten{B}$ for all $i$. The contraction operation can then be written as follows:

\begin{align}
    \ten{C}=\ten{A}\times_{l_1,\cdots,l_h}^{l'_1,\cdots,l'_h} \ten{B}.
\end{align}

\paragraph{Frobenius norm:} The Frobenius norm of a tensor $\ten{A}\in \mathbb{R}^{I_1\times I_2 \cdots \times I_t}$, denoted by $\|\ten{A}\|_F$, is defined as follows:
\begin{align}
\|\ten{A}\|_F=\sqrt{\sum \limits_{i_1,i_2,\cdots, i_t} \left(a_{i_1,i_2,\cdots, i_t}\right)^2 }.
\end{align}

\paragraph{Tensor-Train (TT) Decomposition:}
\label{sec:tensor_train}

For a given tensor $\ten{W}$ and an error bound $\epsilon=\frac{\lVert\ten{W}-\hat{\ten{W}}\rVert_F}{\lVert \ten{W}\rVert_F}$, the TT factors $\{ \ten{G}_j \}_{j=1}^t$, are computed by conducting a hierarchical $\sigma$-truncated singular value decomposition (SVD) for unfolded tensor $\ten{W}$ along its different dimensions, where $\sigma = \frac{\epsilon}{\sqrt{t-1}}\Vert \ten{W} \Vert_F$~\citep{Oseledets-tt-2011}. 

\begin{algorithm}[t]
\caption{TT Contraction Network}
\label{alg:tt-net}
\begin{algorithmic}
\REQUIRE{$\ten{X},\ten{G}_1,\cdots,\ten{G}_{k+d}$} 
\STATE $\ten{Y}=\ten{X}\times_{1}^{2}\ten{G}_1$
\FOR{$t=2$ \textbf{to} $t=k$}
\STATE $\ten{Y}=\ten{Y}\times_{1,4}^{2,1}\ten{G}_t$ 
\ENDFOR
\FOR{$t=1$ \textbf{to} $t=d$}
\STATE $\ten{Y}=\ten{Y}\times_{t+1}^{1}\ten{G}_{k+t}$ 
\ENDFOR
\STATE Reshape $\ten{Y} \in \mathbb{R}^{1\times m_1 \times \cdots \times m_d \times 1}$ to $\mat{y}\in \mathbb{R}^{M}$
\STATE \textbf{Return}  $\mat{y}$

\end{algorithmic}
\end{algorithm}

\section{Tensor Geometry Optimization}
\label{sec:tensor-shape}
To construct a low-rank tensorized layer, a weight $\mat{W}$ is folded to a $(k+d)$ dimensional tensor $\ten{W}$. The memory requirement of a tensorized layer depends on the tensor geometry (shape) selected in the folding process~\citep{mu2013square, Zhong2019, solgi-ts-2023}. It is necessary to find an optimal tensor geometry that minimizes the memory requirement of a low-rank tensorized layer in the TT format as follows:

\begin{align}
    \label{eq:tt_opt_problem}
    \min_{\mat{s}}  & \sum_{j=1}^{k+d} r_{j-1}^{(\mat{s})}\times s_j \times r_j^{(\mat{s})},\\
    {\rm subject \; to}\; & \prod_{j=1}^ks_j=N,
    \prod_{j=k+1}^{k+d}s_j=M,\nonumber\\
    &s_j \geq 2, 
    s_j \in \mathbb{Z},  j=1,\cdots,k+d, \nonumber
\end{align}
where $\mat{s}=(n_1,\cdots,n_k,m_1,\cdots,m_d)$ and $r_j^{(\mat{s})}$ is the TT rank associated with the shape $\mat{s}$. When the ranks are unknown, the above problem is hard to solve. It has been shown that $\min_{\mat{s}}\sum_{j=1}^{k+d} s_j$ serves as an upper bound surrogate for the cost function of Eq.~\eqref{eq:tt_opt_problem} for all ranks $r_j^{(\mat{s})} \in \mathbb{R}$. This upper bound surrogate leads to two independent integer programming problems of minimization of sum under product constraint whose solutions are the feasible and most balanced shapes for the input ($n_1,\cdots,n_k$) and output dimensions ($m_1,\cdots,m_d$), separately~\citep{solgi-thesis-2024}.

\section{DistilBERT Compression Results}
\label{app:DistilBert}
\subsection{Word Embedding Compression}

When the embedding layer is compressed using TT decomposition, the approximation error associated with high-frequency tokens in a sequence accumulates, significantly perturbing the input to the encoder network—even if the individual errors are small. Consequently, incorporating a sparse component for highly frequent tokens helps mitigate error accumulation and the resulting perturbations.

\begin{table}
\small
\caption{Experimental results demonstrate the compression of the word embedding layer of DistilBERT using Saten(e) and TT(e) ($\gamma_{emb}$ denotes the compression factor of the embedding layer, defined as the ratio of the compressed size to the uncompressed size).}
\label{tab:embed_only}
\begin{center}
\begin{tabular}{cc|cc}
\multicolumn{1}{c}{Model}
&\multicolumn{1}{c}{$\gamma_{emb}$}
&\multicolumn{1}{c}{SST2 (Acc\%)}
&\multicolumn{1}{c}{MRPC (F1\%)}
\\ \hline
Base&1.00&91.28&90.22\\
\hline
Saten(e)&0.46&91.17&90.03\\
\hline
TT(e)&0.43&88.18&81.97\\
TT(e)&0.58&90.37&84.97\\
\hline
\end{tabular}
\end{center}
\end{table}

Table~\ref{tab:embed_only} lists the scores and compression factor of the embedding layer denoted by $\gamma_{emb}$ for different datasets and different networks. In Table~\ref{tab:embed_only}, base network refers to the uncompressed DistilBERT network, Saten(e) refers to a network in which only the word embedding layer is replaced by the Saten layer, while the rest of the network (encoder) remains uncompressed. For Saten(e) we set row sparsity to keep only 1,000 tokens resulting in about 96.7\% sparsity ($\rho=0.033$). TT(e) refers to the network whose word embedding layer is compressed with TT without sparsity while the rest of the network remains uncompressed. For TT(e) the low-rank budget has been changed to study the effect of low-rank parameters. Both Saten(e) and TT(e) have been fine-tuned with the same training arguments. 

As shown in Table~\ref{tab:embed_only}, Saten(e) compresses the embedding layer by more than two times without a significant reduction in the evaluation metrics. However, TT(e) shows a significant drop in evaluation metrics while having almost the same compression efficiency. TT(e) with less memory reduction compared to that of Saten(e), has worse evaluation metrics, emphasizing that sparsity played a crucial role. Therefore, simply increasing the rank of the TT does not provide a matching evaluation metric to that of Saten(e).  

\begin{figure}[!t]
\centering
\includegraphics[width=3in]{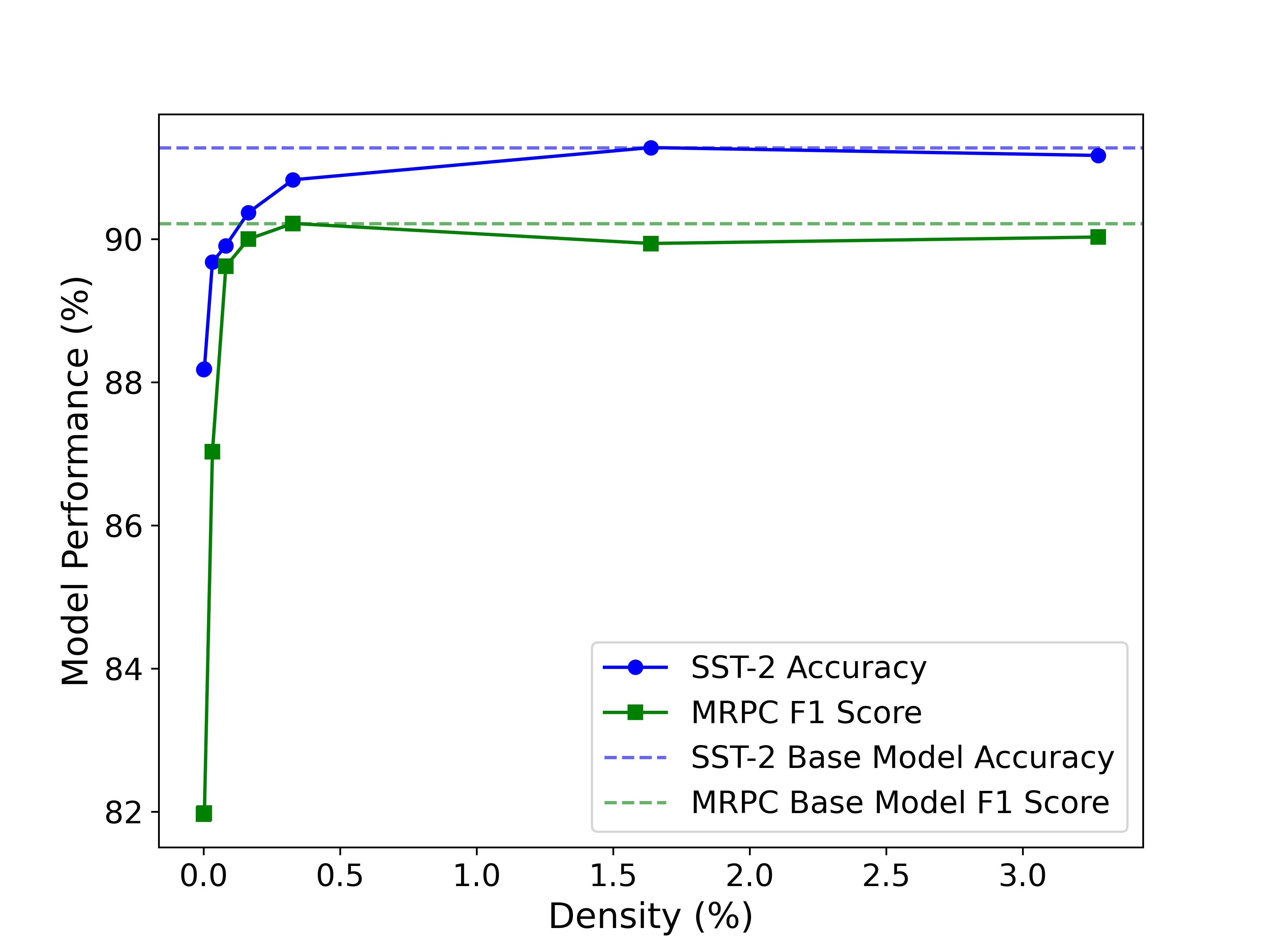}
\caption{Accuracy versus density of saten(e) for SST2 and MRPC datasets.}
\label{fig:density_accuracy}
\end{figure}

Fig.~\ref{fig:density_accuracy} illustrates the accuracy versus the density of the sparse component of the saten(e) for SST2. It is observed that even a high sparsity significantly enhances the network's accuracy. For instance, from TT only (without sparsity component) to Saten(e) with $99.84\%$ sparsity ($\rho=0.0016$, corresponding to only the 50 most common tokens), the accuracy of the network increases from 88.18 to 90.37.

\subsection{Low-Rank Versus Sparsity}

\begin{table*}[h]
\small
\caption{The experimental results of full model compression of DistilBERT on the SST-2 and MRPC datasets using Saten(u), Saten(2:4), and TT.}
\label{tab:distilbert}
\begin{center}
\begin{tabular}{ccc|ccc}
\multicolumn{1}{c}{Model}
&\multicolumn{1}{c}{$L_{lin}$ (\%)}
&\multicolumn{1}{c}{$\rho_{lin}$ (\%)}
&\multicolumn{1}{c}{Score (\%)}
&\multicolumn{1}{c}{\#Params (M)}
&\multicolumn{1}{c}{\#MACs (M)}
\\ \hline
DistilBERT-SST2&-&-&91.28&67.5&43.1\\
\hline
Saten(u)&31&50&91.51&47.6&38.6\\
Saten(u)&31&20&90.94&34.4&25.4\\
Saten(u)&31&10&89.11&28.7&19.7\\
\hline
Saten(2:4)&20&50&91.63&41.6&32.2\\
Saten(2:4)&8&50&90.83&36.1&25.9\\
Saten(2:4)&0&50&89.44&32.8&21.5\\
\hline
TT&76&0&90.02&43.8&35.5\\
TT&31&0&82.22&24.1&15.9\\
\hline
\hline
DistilBERT-MRPC&-&-&90.22&67.5&43.1\\
\hline
Saten(u)&31&25&88.48&35.4&26.4\\
Saten(u)&31&20&88.04&33.2&24.2\\
Saten(u)&31&10&85.11&29.9&20.9\\
\hline
Saten(2:4)&20&50&88.40&41.6&32.2\\
Saten(2:4)&8&50&87.67&36.1&25.9\\
Saten(2:4)&0.0&50&82.22&32.8&21.5\\
\hline
TT&76&0&83.75&43.0&35.5\\
TT&31&0&82.21&43.0&35.5\\

\hline
\end{tabular}
\end{center}
\end{table*}

Table~\ref{tab:distilbert} lists the results of end-to-end compression of DistilBERT. In Table~\ref{tab:distilbert}, Saten(u) and Saten(2:4) refer to models in which the encoder network is compressed using unstructured and 2:4 sparsity patterns, respectively, while the word embedding layer is compressed using row sparsity, as in Saten(e). TT refers to a network where the word embedding layer is compressed using Saten, while the encoder is compressed using TT. In Table~\ref{tab:distilbert}, the scores for SST-2 and MRPC refer to Accuracy and F1-Score, respectively.

For Saten(u) the low-rank budget is kept the same, but the sparse budget has been changed to show the effect of sparsity on Saten versus the model accuracy. On the other hand, for Saten(2:4) the sparsity budget is constant due to the pattern of sparsity, but the low-rank budget has been changed to investigate the effect of low-rank component on the network's performance score. Note that for Saten(2:4) with $L_{lin}=0$, although the low-rank budget is set to zero, the sparse pattern has been derived based on the low-rank decomposition error. For all Saten networks, the total embedding parameters are 11.3 million with near 97\% sparsity for the word embedding layer. 

For SST2 dataset we can observe that both Saten(u) and Saten(2:4) were able to compress the network by 30\% and 40\%, resepctively with no drop in accuracy. For both MRPC and SST2 we were able to compress the DistilBERT by more than two times with at most 2\% drop in the network's score. Considering the fact that DistilBERT network is already a compressed model, this level of compression with no significant drop of accuracy is notable and shows the capacity of the Saten framework for compressing language models.

Comparing the TT and Saten models for both SST2 and MRPC, TT leads to a more significant drop in network's score than Saten even with almost the same or worse compression ratios. Comparing TT and Saten(u) that have the same low-rank budget, we can observe that by adding a small sparse budget the network achieves significantly higher scores for both MRPC and SST2 dataset. On the other hand, by increasing the low-rank budget for TT, we do not observe the same amount of improvement.

Studying the effect of sparsity and the low-rank component, it is evident that both play a key role in preserving the performance of the models. Comparing Saten(u) and Saten(2:4) with TT, we can observe that the Saten models not only improve the compression efficiency but also achieve higher scores. 

\section{Implementation and Fine-Tuning Details}
\label{app:run-details}
For our experiments we downloaded the pre-trained models and data from Huggingface~\citep{wolf-transformers-2020}.  The BERT models are under Apache 2.0 license. In this we also applied LLaMA 3.2-1B model, developed by Meta Platforms, Inc., available under the LLaMA 3.2 Community License. For all applied datasets, we used all training data for fine-tuning and we evaluated all models using all the evaluation datasets. All reported evaluation metrics are one-time run. For fine-tuning of the base models we applied the default setting of the downloaded models. For both BERT-Base and LlaMA models we used learning rate 2e-5 and 5 epochs. We used batch size 8 and 2 for BERT-Base and LlaMA models, respectively. We fine tuned the models on single NVIDIA RTX 6000 ADA generation GPU using the default AdamW optimizer with weight decay of 0.01. After compressing the models we fine-tuned for another 5 epochs with the same settings but we used learning rate 5e-6 for compressed BERT-Base models and learning rate 2e-5 (2e-6 for BoolQ) for compressed LlaMA-3.2-1B models. 

For compressing the models using Saten we need to set TT error bound $\epsilon$ and sparsity ratio for each layer in the network. For BERT-Base experiments for Saten(u) we set $\epsilon=0.75$ for all linear layers and for Saten(2:4) we used $\epsilon=1$ for all linear layers.  For LlaMA experiments, Saten(u) uses an error bound $\epsilon=1$ for all linear layers but for query and key layers we used $\epsilon=0.4$. For Saten(2:4) we used $\epsilon=1$ for all layers but for query and key layers we used $\epsilon=0.65$. For embedding layers for both BERT-Base and LlaMA we applied $\epsilon=0.5$ and used 1,000 ($\rho=0.03$) and 10,000 ($\rho=0.08$) tokens for sparsity, respectively. For the sparsity of Saten(u) in all experiments of BERT-Base and LlaMA-3.2-1B, we set the sparsity to be 95\% for all linear layers, but 90\% for query and key layers, which on average is about 95\% sparsity given the smaller size of query and key layers. 

For TT compression of BERT-Base we used $\epsilon=0.65$ for all compressed linear layers. We did not compress the query and key linear layers as it degraded its performance. In LlaMA-3.2-1B experiments we run TT twice with different budgets to compare with both Saten(u) and Saten(2:4). For comparison with Saten(u) we used $\epsilon=0.4$ for query and key layers and $\epsilon=0.9$ for the rest of linear layers. This setting was set to be similar to Saten(u) but gives a higher number of parameters to TT. For comparison with Saten(m:n) we set $\epsilon=0.6$ and $\epsilon=0.55$ for the query/key linear layers and the rest of the linear layers, respectively. For Tucker decomposition, we applied similar setting as the TT but to match the compression ratio with Saten(u) we set $\epsilon=1.0$ for all layers. 

\section{Llama-3.2-3B}
\begin{table}[t]
\small
\caption{Results of compressing LlaMA-3.2-3B using Saten versus SVD and TT for WikiText-2 dataset (CR denotes compression ratio).} 
\label{tab:llama3}
\begin{center}
\begin{tabular}{c|ccccc}
\multicolumn{1}{c}{Model}
&\multicolumn{1}{c}{Zero-Shot (PPL)}
&\multicolumn{1}{c}{LoRA (PPL)}
&\multicolumn{1}{c}{CR}
\\ \hline
Base&7.60&7.18&1.0\\
Saten(2:4)&780.66&13.37&0.6\\
TT&312755.56&1813.55&0.6\\
SVD&120484.91&776.16&0.6\\
\hline
\end{tabular}
\end{center}
\vspace{-10pt}
\end{table}

Additionally, we compressed Llama-3.2-3B for Wikitext-2 dataset \citep{wikitext}. To ensure the generalization of our approach to larger models, we directly compressed the pre-trained model without initial fine-tuning, followed by parameter-efficient fine-tuning using low-rank adapters (LoRA), which is standard practice for large-scale models~\citep{hu2021lora}. We also reported perplexity as a proxy for the model evaluation across different tasks. The results presented in Table~\ref{tab:llama3} show that Saten significantly improves perplexity (PPL).

\end{document}